# A neural network based heading and position control system of a ship

Shahroz Unar [a,*], Mukhtiar Ali Unar [b], Zubair Ahmed Memon [c], Sanam Narejo [b]

[a] *The Institute of Information & Communication Technologies, Mehran University of Engineering & Technology, Jamshoro Pakistan*

[b] *Department of Computer Systems Engineering, Mehran University of Engineering and Technology, Jamshoro Pakistan*

[c] *Department of Electrical Engineering, Mehran University of Engineering and Technology, Jamshoro Pakistan*

* Corresponding author: Shahroz Unar, Email: shahroz.unar@gmail.com

A B S T R A C T

Heading and position control system of ships has remained a challenging control problem. It is a nonlinear multiple input multiple output system. Moreover, the dynamics of the system vary with operating as well as environmental conditions. Conventionally, simple Proportional Integral Derivative controller is used which has well known limitations. Other conventional control techniques have also been investigated but they require an accurate mathematical model of a ship. Unfortunately, accuracy of mathematical models is very difficult to achieve. During the past few decades computational intelligence techniques such as artificial neural networks have been very successful when an accurate mathematical model is not available. Therefore, in this paper, an artificial neural network controller is proposed for heading and position control system. For simulation purposes, a mathematical model with four effective thrusters have been chosen to test the performance of the proposed controller. The final closed loop system has been analysed and tested through simulation studies. The results are very encouraging.

## 1. Introduction

Control of surface ships has always remained a challenging problem. Ship control systems are being developed for more than a century but research on better controller design continues. The International Federation of Automatic Control (IFAC) considers the ship control problem among one of its benchmarking problems because it is a highly non-linear control problem. Moreover, the ship dynamics vary with operating as well as environmental conditions. The operating conditions include the speed of the vessel, the loading conditions and trim etc. On the other hand, the environmental conditions include the wind generated waves, the depth of water and the ocean currents. The design of a robust controller under all these conditions is not an easy task. During the past hundred years, almost all conventional controller techniques such the Proportional Integral Derivative (PID), Pole Placement, the Least Quadratic Gaussian (LQG), Feedback Linearization, Sliding Mode Control, Back Stepping Control and various adaptive techniques have been explored. A comprehensive review of the mentioned conventional techniques and their references can be found in [1]. Unfortunately, every technique suffers from one or another problem and it has not been possible to design a controller which is free from all disadvantages.

One common problem in almost all of the above control techniques is that an accurate mathematical model is required to tune the parameters of the controller. A ship is generally modelled as a 6 Degree of Freedom (DoF) non-linear model representing three linear and three angular motions. A detailed derivation of these equations can be found elsewhere [2].

Neural Network based techniques may be suitable for course control system because they do not require mathematical model of the vessel for controller design. A neural network based control system may be developed which may behave optimally at varying operating and environmental conditions without compromising stability boundaries.

In this paper, a neural network based controller is proposed which controls heading and position of a ship. This is done by controlling the azimuth angles of four thrusters.

## 2. Ship Course Control Problem

In a ship course control system, controller is designed so that the ship follows a desired course. This is called course keeping control problem. On the other hand, a controller may be designed so that the ship turns at a desired heading angle. This is termed to as the course changing problem. In both of these control systems, usually rudder motion is controlled to achieve a desired heading during the course keeping or course changing mode. However, in some modern ships, effective thrusters are used instead of a rudder. Thruster based control system is particularly suitable in position keeping applications.

A typical course control system is shown in Fig. 1, in which $\Psi_r$, $\Psi_d$ and $\Psi$ are the reference heading, desired heading and actual heading respectively. $\delta_c$ is the commanded rudder angle and $\delta$ is the actual rudder angle. This simple control system is also called a ship autopilot where only rudder angle is controlled to achieve the desired course. Sometimes, roll motion is also controlled along with the heading angle. This is called the rudder roll stabilization. However, this aspect will not be considered in this paper.

In Fig. 1, reference model is used to generate the desired heading angle. For a course changing maneuver, a second order transfer function of the following form may be used.

$$\frac{\Psi_d}{\Psi_r} = \frac{\omega_n^2}{s^2+2\zeta\omega_n s+\omega_n^2} \qquad (1)$$

where $\omega_n$ and $\zeta$ are un-damped natural frequency and damping ratio respectively. The damping ratio should be unity or near unity for a critically damped system. The steering machine model is shown in Fig, 2. This model of the steering machine was first proposed by Van Amerongen [3].

The controller of Fig. 1 is also called autopilot because it automatically controls the heading of the ship under consideration under varying operating and environmental conditions. In case of thruster driven ship, the effective thrusters are used instead of a rudder. In general, the rudder based control system is a Single Input Single Output (SISO) control system whereas the thruster based system is a multivariable control system because more than one thrusters are used to control heading and position of a ship.

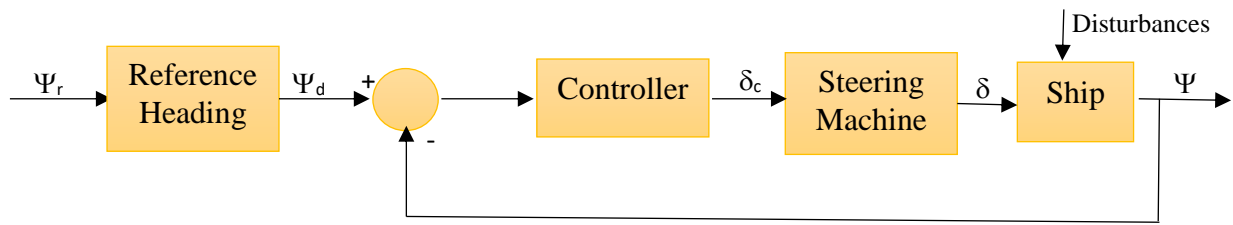

**Fig. 1.** Ship steering control system

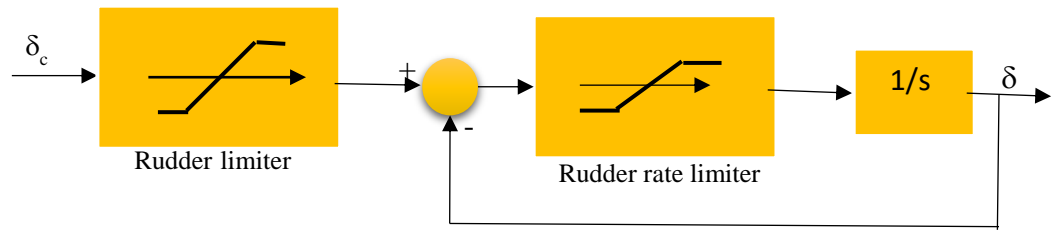

**Fig. 2.** Steering machine

## 3. Literature Review

In the past, artificial neural networks have been extensively used for both for course keeping and course changing manoeuvres. Richter and Burns [4], Witt et al. [5], Burns [6] and Simensen and Murray-Smith [7] were among the first who explored the applicability of artificial neural networks for heading control of a ship. They developed neural network based controller in such a way that it behaved like a conventional PID controller. They trained the controller by using the conventional error back propagation algorithm. Unar and Murray-Smith [1, 8] introduced the applicability of Radial Basis Function networks for the application. After that pioneering work, a lot of neural network based control systems have been proposed. Most recent work include Tung [9], Haouari et al. [10], Wang et al. [11], and Wang [12].

Due to space limitation, it is not possible to present a comprehensive literature review. However, it can easily be concluded as follows.

1. In most of the previous work, the feedforward neural networks (either Multilayer Perceptron (MLP) networks or Radial Basis Function (RBF) networks) have been used.

2. In almost all closed loop control systems, a rudder angle has been controlled for course keeping or course changing or both control systems.

3. Very little work has been done on thruster driven ships.

Almost all of the researchers have applied the feedforward neural networks (either Multilayer Perceptron Networks or Radial Basis Function Networks) for the application. Moreover, the controlling variable in all of the above mentioned papers is the rudder angle.

This paper proposes a neural network based controller for a thruster driven ship. As mentioned, thruster based control system is more challenging because more than one thrusters are to be controlled. In other words, the thruster based control system is a Multiple Input Multiple Output (MIMO) control system.

## 4. Thruster Based Control System

Thruster based control systems are very effective in Dynamic Positioning (DP) of ships. A thruster based control system is able to maintain a predetermined position and heading automatically by means of thruster force.

The advantages of thruster based control systems are many and include the following [13, 14].

1. Enhance the manoeuvrability of the vessel especially at low speeds

2. Less sensitive to vibration and noise

3. Make docking easier, since they allow the captain to turn the vessel to port or starboard side, without using the main propulsion mechanism which requires some forward motion for turning.

4. Provide better electrical efficiency, better use of ship space and better hydrodynamic efficiency.

5. Lower maintenance cost

## 5. Mathematical Model

The mathematical model chosen has been taken from McGookin et al. [15]. This is a scale model of an experimental ship available at the Engineering Cybernetics Department of Trondheim University Norway. The ship is titled as Cyber-Ship-I. The length of this scale is 1.17 m. This is $1/70^{th}$ the size of the actual ship. The linear as well as angular velocities of Cyber-Ship-I are approximately eight times larger than the actual vessel's [15]. This ship has four thrusters, two at the bow and two at the stern, as shown in Fig. 3. These thrusters are used to control heading and position of the ship. The surface motion of this ship can be represented by Eq. 2.

$$M\dot{v} + C(v)v + D = \tau \qquad (2)$$

where M, C and D are the mass/inertia, Coriolis and damping matrices respectively.

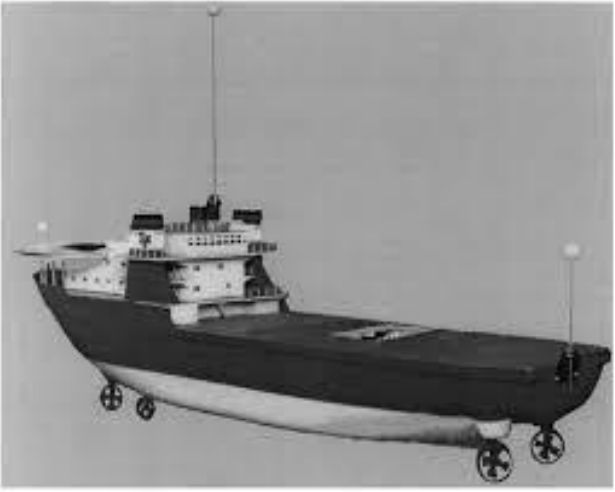

**Fig. 3.** Cyber-Ship-I

The variable v in Eq. 1 is the vector $\upsilon = [u \; v \; r]$ where u is the surge velocity in m/s, v is the sway velocity in m/s, and r is the heading rate in degrees/s. The variable $\tau$ is the input force vector provided by the thrusters. It is a vector $\tau = [\tau_x \; \tau_y \; \tau_z]$, where $\tau_x$ is the thrust force along the body fixed x-axis, $\tau_y$ is the thrust force along the body fixed y-axis and $\tau_z$ is the thrust force along the body fixed z-axis.

For the ship under consideration, the matrices M, C and D have been computed as follows [15].

$$M = \begin{bmatrix} 19 & 0 & 0 \\ 0 & 35.2 & 0 \\ 0 & 0 & 20 \end{bmatrix},$$

$$C(\upsilon) = \begin{bmatrix} 0 & 0 & -35.2v \\ 0 & 0 & 19u \\ 35.2v & -19u & 0 \end{bmatrix}$$

and $D = \begin{bmatrix} 6.3 & 0 & 0 \\ 0 & 7 & 0 \\ 0 & 0 & 2 \end{bmatrix}$

Eq. 2 may also be re-written as Eq. 3.

$$\dot{\upsilon} = -M^{-1}(C(\upsilon)\upsilon + D) + M^{-1}\tau \qquad (3)$$

The Kinematics equations are mathematically represented as Eq. 4.

$$\dot{\eta} = J(\eta)\upsilon \qquad (4)$$

where $J(\eta) = \begin{bmatrix} \cos\Psi & -\sin\Psi & 0 \\ \sin\Psi & \cos\Psi & 0 \\ 0 & 0 & r \end{bmatrix}$ denotes Euler equations relating the body fixed and the Earth fixed reference frames and $\eta = [x \; y \; \Psi]$, where $\Psi$ is the yaw (heading) angle in degrees and (x, y) is the x and y position of the ship.

In the standard state space representation, Eqs. (2) and (3) can be re-arranged as Eq. 5.

$$\dot{x} = Ax + B\tau \qquad (5)$$

where $\dot{x} = \begin{bmatrix} \dot{\upsilon} \\ \dot{\eta} \end{bmatrix}, x = \begin{bmatrix} \upsilon \\ \eta \end{bmatrix},$

$$A = \begin{bmatrix} -M^{-1}(C(\upsilon))\upsilon + D) & 0 \\ J(\eta) & 0 \end{bmatrix}, B = \begin{bmatrix} M^{-1} \\ 0 \end{bmatrix}$$

The thruster configuration for Cyber-Ship-I is shown in Fig 4. In the Figure, the position of the thrusters is given relative to the center of gravity which is also the origin of the body fixed reference frame. Each thruster is represented by (a) the force it produces (i. e. $f_i$; i=1, 2, 3, 4) and (b) the azimuth angle $a_i$ defining the direction of the corresponding force, as shown in Fig. 2. The azimuth angles $\alpha_1$ and $\alpha_2$ can be set independently, however, $\alpha_3$ must be equal to $\alpha_4$ so that the thruster 3 and thruster 4 operate in the same direction.

The thruster forces and azimuth angles are mathematically related to each other as Eq. 6.

$$\tau = H(\alpha)f \qquad (6)$$

where

$f = [f_1, f_2, f_3, f_4]^T, \alpha = [\alpha_1, \alpha_2, \alpha_3, \alpha_4]^T$ and

$H(\alpha) =$
$$\begin{bmatrix} \cos\alpha_1 & \cos\alpha_2 & \cos\alpha_3 & \cos\alpha_4 \\ \sin\alpha_1 & \sin\alpha_2 & \sin\alpha_3 & \sin\alpha_4 \\ 0.497\sin(\alpha_1-\theta_1) & 0.497\sin(\alpha_2-\theta_2) & 0.407\sin\alpha_3 & 0.527\sin\alpha_3 \end{bmatrix}$$
...(7)

where $\theta_1$ and $\theta_2$ are the phase shift angles of the thrusters relative to the center of gravity. Due to their very small values, these phase shift angles are assumed to be zero in this paper.

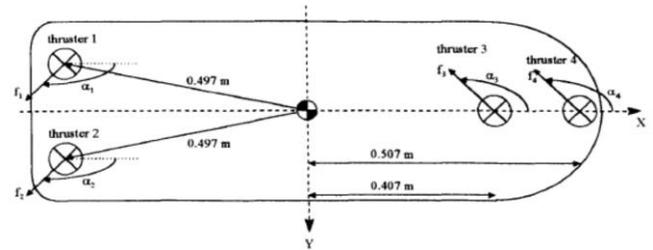

**Fig. 4.** Thruster configuration

In the simulation studies, following values of $\alpha_1$, $\alpha_2$, $\alpha_3$ and $\alpha_4$ have been used as suggested by McGookin [15].

$\alpha_1 = \alpha_2 = \pi$ radians and $\alpha_3 = \alpha_4 = \pi/2$ radians

Substituting the above values of the azimuth angles and ignoring phase shifts, we get Eq. 8.

$$H(\alpha) = \begin{bmatrix} -1 & -1 & 0 & 0 \\ 0 & 0 & 1 & 1 \\ 0 & 0 & 0.407 & 0.527 \end{bmatrix} \qquad (8)$$

This indicates that the surge motion is governed by thrusters 1 and 2 at the stern and the sway and yaw motion is governed by thrusters 3 and 4.

## 6. Results and Discussion

A multilayer Perceptron neural network with an input layer, hidden layer and output layer is trained with 10 neurons in the hidden layer. Because such a network is trained by using the supervised learning, so a well optimized sliding mode controller is used for generating the desired data for training. The heading response should be critically damped with zero overshoot.

1. After turning at a particular angle, the ship should follow a path parallel to the original path.

2. The difference between the actual and the desired heading should be as small as possible.

3. The Heading angle rate should be smooth and should approach towards zero after turning.

4. The difference between the actual and desired heading should be minimum

Fig. 5 shows heading response for a step command of 20°. The response is very smooth and critically damped. After turning, the ship's heading is parallel to the actual path. The heading error is plotted in Fig. 6. For the given ship, the maximum allowable heading angle error is ±1°, however, the maximum heading error generated by the proposed controller is approximately 0.4° which is well within the acceptable limits. It can easily be seen that the steady-state error is zero. The heading rate is depicted in Fig. 7. The heading rate error is less than 1° and its steady state value is zero. It is therefore concluded that the heading response is highly satisfactory.

The graph between x-position and y-position is shown in Fig. 8. It is desired that none of the control signals which reach saturation. The smooth responses indicate that none of the three control signals saturate.

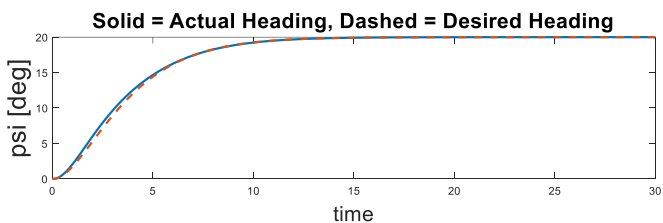

**Fig. 5.** Actual heading and desired heading when the vessel turns at an angle of 20°

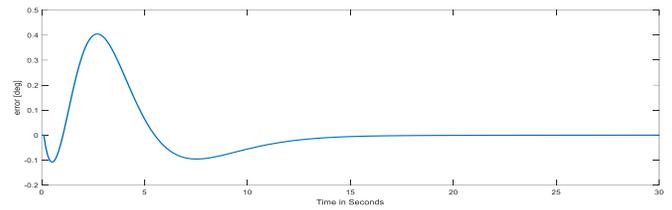

**Fig. 6.** Heading error in degree

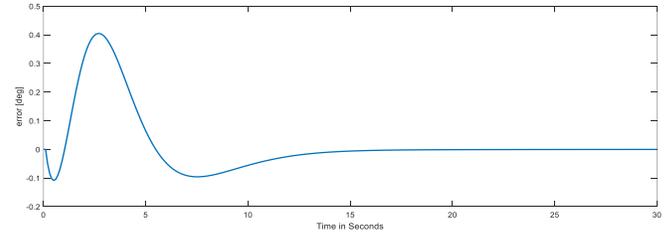

**Fig. 7.** Heading error rate in degrees

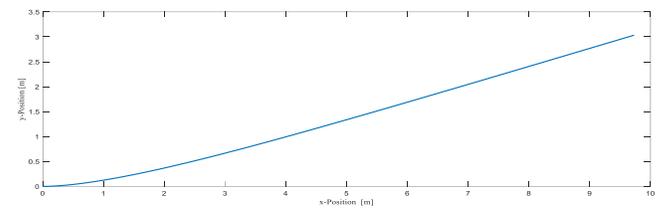

**Fig. 8.** x and y position

## 7. Conclusion

This paper proposes an artificial neural network based controller for heading and position control of a ship. A mathematical model of a ship has been derived from the physical model of Cyber-Ship-I, which is an experimental facility available at University of Trondheim Norway. The simulation results show that the proposed controller has produced satisfactory results. The controller has controlled heading and position of the vessel with reasonable accuracy. In the future work, the performance of the proposed controller may be investigated under external disturbances.